\documentclass{article}

\usepackage[final,nonatbib]{icbinb_neurips_2020}

\usepackage[utf8]{inputenc} 
\usepackage[T1]{fontenc}    
\usepackage{hyperref}       
\usepackage{url}            
\usepackage{booktabs}       
\usepackage{amsfonts}       
\usepackage{nicefrac}       
\usepackage{microtype}      

\usepackage{amsmath}
\usepackage{graphicx}

\usepackage{color}

\usepackage{enumitem}

\title{Adversarial training for predictive tasks:
theoretical analysis and limitations in the deterministic case}

\author{Thibault Lesieur, J\'er\'emie Messud, Issa Hammoud, Hanyuan Peng,\\
\textbf{C\'eline Lacombe, Paulien Jeunesse}\\
CGG, 
Subsurface Imaging R\&D, Massy (France)
}
\begin{document}

\maketitle

\begin{abstract}
To train a deep neural network to mimic the outcomes of processing sequences,
a version of Conditional Generalized Adversarial Network (CGAN) can be used.
It has been observed by others that CGAN can help to improve the results
even for deterministic sequences, where only one output is associated with the processing of an input.
Surprisingly, our CGAN-based tests
on some deterministic geophysical processing sequences did not produce a real improvement compared to the use of an $L_p$ loss;
we here propose a first theoretical explanation why.
Our analysis goes from the non-deterministic case to the deterministic one.
It led us to develop an adversarial way to train a content loss that gave better results on our data.
\end{abstract}


\section{Introduction}

\begin{figure}[ht]
\centering
\includegraphics[width=1.00\linewidth]{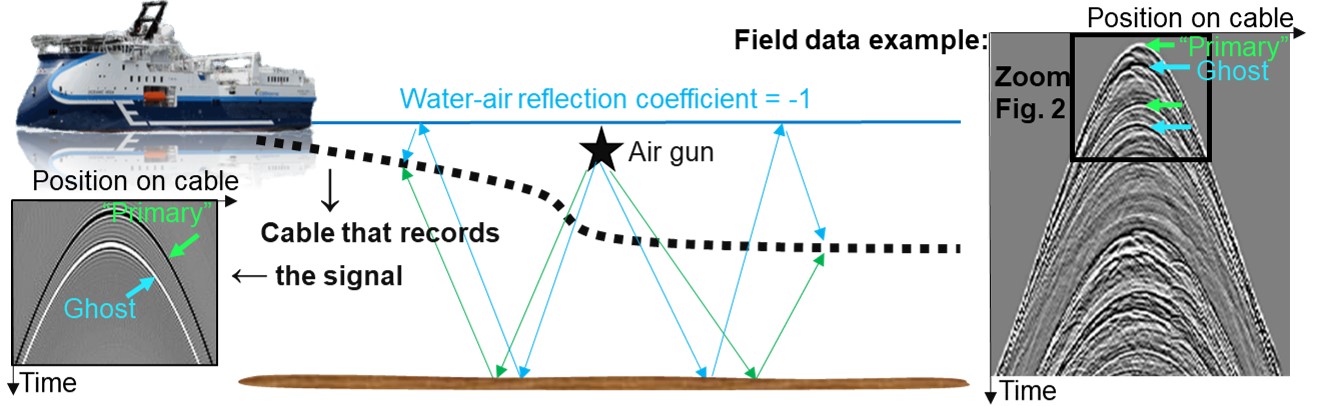}
\caption{
Marine seismic data acquisition.
The pressure wavefield generated by an airgun is reflected in the subsurface, then comes back to the surface
and is recorded along a cable pulled by a boat.
Billions of data over thousands of square kilometers are recorded.
A particularity of the geophysical data is to consist of very structured and continuous events, corresponding to 
discontinuities (layers) in the subsurface.
The greyscale represents the polarity of the wavefield (black: positive, white: negative).
Blue highlighted events have reflected on the water surface and are called ``ghosts'';
they look like ``duplicate'' events with reverse polarity;
they interfere with the other events
and must be removed by a ``deghosting'' sequence for some further applications.
}
\label{fig:figure1}
\end{figure}

We consider the problem of mimicking a complicated processing sequence
by learning some representation of the joint probability density function (pdf)
that couples the outcomes of the sequence to its inputs.
%
%
In geophysics, our field of application, processing sequences are usually based on workflows that represent a combination of algorithms and user-provided information to achieve a given task.
Wave equation and signal processing are classical components of the algorithms,
and geological priors are often part of user-provided information
\cite{Yilmaz2001}.
Several geophysical processing sequences aim at removing some 
undesired very structured events in the geophysical data \cite{Yilmaz2001},
like the ``ghost'' events illustrated in Fig.~\ref{fig:figure1}.
Learning an efficient representation that mimics such sequences can bring value,
for example to take the best of various existing workflows, increase turnaround or obtain a processing guide.
Deep Neural Networks (DNNs) provide a flexible tool to parameterize a function that predicts outcomes from inputs.
Many explorations have recently been done using DNNs to mimic geophysical processing sequences,
see for instance Refs.
\cite{Alwon2018, Picetti2018, Halpert2018, Si2019, Zhang2020}.

To train a DNN to predict outcomes from inputs,
we may consider methods inspired from
the Generative Adversarial Network (GAN) framework \cite{goodfellow2014generative},
in particular Conditional GAN (CGAN) \cite{Mirza2014,Isola2017}.
Indeed, CGAN can deal with joint pdfs
(contrary to the original GAN formulation that deals only with single parameter pdfs),
the originality being that the discriminator becomes conditioned by the input data \cite{Ledig2017,Halpert2018,Picetti2018,Zhang2020}.
However, in the common context of a deterministic processing sequence,
where only one outcome is generated when the sequence is applied to an input \cite{Alwon2018, Picetti2018, Halpert2018, Si2019, Zhang2020},
using a simple $L_p$ norm based loss for the training usually gives good results \cite{Goodfellow2016}.
So, can CGAN be pertinent even in the deterministic case?
It has been observed that combining CGAN with an $L_p$ loss may help to improve the results further,
see e.g.
Ref. \cite{Ledig2017} for natural image processing and
Refs. \cite{Halpert2018,Zhang2020} for geophysical processing.

\begin{figure}[ht]
\centering
\includegraphics[width=1.0\linewidth]{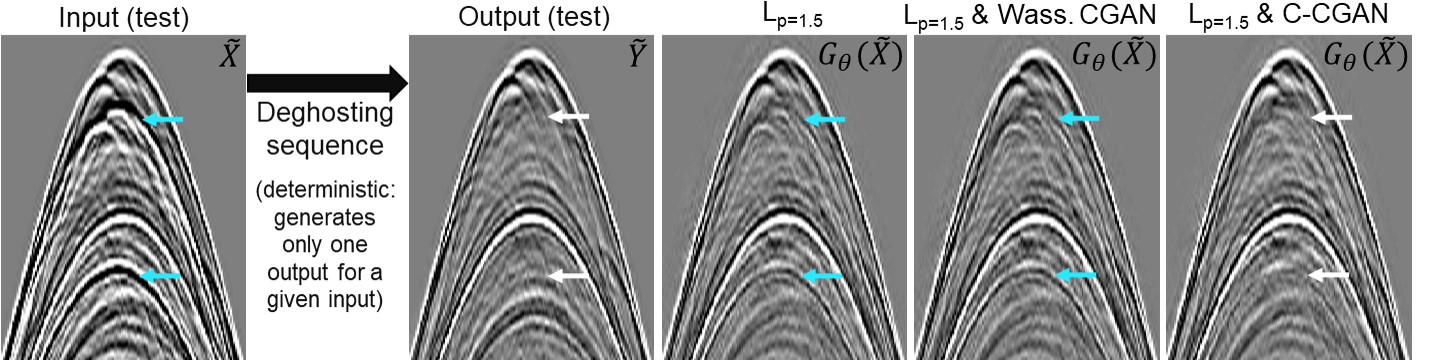}
\caption{
The DNN training data consists in 200 randomly extracted input ``images'' of size $564 \times 551 \times 1$,
representing 0.001\% of the total field data, together with corresponding output images generated by a conventional deghosting sequence.
On the left, a conventional deghosting result is shown on a test data (chosen ``far'' from the training data).
On the right, various DNN predictions from the input test data, shown before training full convergence to highlight the differences (40 epochs training).
Ghost residuals (blue arrows) can be observed using an $L_p$ loss and adding W-CGAN does not improve the result.
Adding our C-CGAN more satisfyingly removes the ghost residuals (white arrows).
However, the main benefit of C-CGAN seems here to accelerate the training as at full convergence (150 epochs) the difference between C-CGAN and $L_p$ becomes quite smaller.
}
\label{fig:figure2}
\end{figure}

Surprisingly, our Wasserstein CGAN-based trainings
\cite{Fabbri2017}
on deterministic geophysical processing sequences,
like the ``deghosting'' (or ghost removal) sequence \cite{Wang2013},
did not help to produce a real improvement in our tests compared to the use of an $L_p$ loss, see
Fig.~\ref{fig:figure2}.
In this paper, we propose a theoretical analysis of this aspect.
First, we remind why $L_p$ losses should perform well in the deterministic prediction case.
Then, we point out from the Wasserstein point of view what CGAN should bring compared to an $L_p$ loss,
taking the opportunity to discuss the Wasserstein CGAN (W-CGAN) foundations.
Our analysis gives a first explanation of why CGAN may perform more poorly than expected,
and also leads to a proposal of an adversarial way to train a content loss 
that we call ``Content CGAN'' (C-CGAN); it gave better results on our data as illustrated in Fig.~\ref{fig:figure2}.
For completeness, we start all our theoretical considerations 
from the non-deterministic prediction case, where multiple outcomes related to one input are possible,
and then take the deterministic limit.

\section{Notations}
\label{sec:Notations}

$\mathcal{X}$ denotes the input data (image) space and $\mathcal{Y}$ the output data (image) space.
${P}_{Y,X}=P_X P_{Y|X}$ denotes the joint pdf associated to the (possibly non-deterministic) processing sequence we wish to mimic.
$P_X$ is the marginal pdf that describes the distribution of the input data;
realizations of the random variable 
$X\sim{P}_{X}$ are denoted by $\tilde{X}\in \mathcal{X}$.
$P_{Y|X}$ is the conditional pdf that describes the
outcomes of the processing sequence related to a given input;
realizations of the random variable 
$Y\sim P_{Y|X}$ are denoted by $\tilde{Y}\in \mathcal{Y}$.
%
$G_\theta^Z:\mathcal{X}\rightarrow \mathcal{Y}$
represents a prediction function parameterized by a model $\theta$, here a DNN,
where the latent space random variable $Z\sim P_Z$
gives the flexibility to produce multiple outcomes related to a given input (for the non-deterministic prediction case).
$\theta$ is to be optimized so that the $Z$-realizations of $G_\theta^Z(X)$ tend to mimic the realizations of 
$Y\sim P_{Y|X}$ .
$\mathbb{E}$ denotes the expectation over a specified random variable.

The deterministic prediction limit can be taken considering both:
\begin{itemize}[leftmargin=1cm ,parsep=0cm,itemsep=0cm,topsep=0cm]
\item
The ``empirical'' joint  pdf, for instance, for 
$
{P}_{{Y},{X}}(\tilde Y,\tilde X)
\rightarrow
\frac{1}{N_D}\sum_{i=1}^{N_D} \delta(\tilde Y-\tilde Y_i)\delta(\tilde X-\tilde X_i)
$.
$\{\tilde X_i,\tilde Y_i;i=1..N_D\}$ denotes a set of input and output data realizations.
%
\item
$G_\theta^Z$ independent of $Z$, so that a unique outcome is predicted by the DNN for each input.
\end{itemize}

\section{Which processing sequences are suitable for the use of an \texorpdfstring{$L_p$}{Lp} loss?}
\label{sec:more_det}

$L_p$-based losses are defined for $p\ge 1$ by
\begin{eqnarray}
\label{eq:dist3}
C^p_{}(G_\theta^Z)
&=&
\mathbb{E}_{Z\sim {P}_{Z}}
\mathbb{E}_{(Y,X)\sim {P}_{{Y}, X}}
||Y-G_\theta^Z(X)||_{L_p}^p
,
\end{eqnarray}
where the output image space $L_p$ norm is defined by
\begin{eqnarray}
||\tilde{Y}-\tilde{Y}^{(2)}||_{L_p}^p
=
\int_{\Omega}
\Big|
\tilde{Y}(y)-\tilde{Y}^{(2)}(y)
\Big|^p
d\mu(y)
,
\quad \forall (\tilde{Y},\tilde{Y}^{(2)})\in \mathcal{Y}\times\mathcal{Y}
.
\label{eq:norm_pixels}
\end{eqnarray}
$\mathcal{Y}$ here represents the real $L^p(\Omega)$ space
(functions with integrable moments of order $p$).
Each $\tilde{Y}\in \mathcal{Y}$ represents an image indexed by the positions $y$
in a ``pixels space'' $\Omega$ that is measurable for the measure $\mu(y)$.
$\Omega$ usually represents the (discrete) pixels grid space and $\mu$ the counting measure.
However, our considerations generalize to continuous spaces $\Omega$ taking the Lebesgue measure for $\mu$.

Training aims to minimize $C^p(G_\theta^Z)$, eq. (\ref{eq:dist3}), with respect to $\theta$.
As $C^p$ measures a similarity between one realization of ${Y}$ and one realization of $G^Z_\theta(X)$,
the trained prediction function $G_\theta^Z$ will tend to become independent of $Z$ and
output some ``average'' 
of all outcomes related to one input data.
Indeed, we can easily compute the optimum for $p=2$:
$G_\theta^Z(X)\approx\mathbb{E}_{Y\sim {P}_{{Y}|{X}}}Y$, and for $p=1$:
$G_\theta^Z(X)\approx\mathbb{M}_{Y\sim {P}_{{Y}|{X}}}Y$
where $\mathbb{M}$ denotes the median
(see Ref. \cite{Goodfellow2016} section 6.2.1.2).

In the non-deterministic prediction case, if the multiple outcomes are 
related to structured events,
training with an $L_p$ loss is obviously not suitable as it would tend to produce blurry predictions (due to the ``averaging'').
However, if the multiple outcomes are related to zero-``average'' noise,
an $L_p$ loss is suitable and would even tend to produce denoised predictions.
Of course, an $L_p$ loss is also suitable to the deterministic prediction case;
a denoising effect can still occur if each of the single outcomes are affected by zero-``average'' noise.
Note that $p=1$ (the median) is more robust to outliers than $p=2$ but harder to train.
$p=1.5$ has been chosen in Fig.~\ref{fig:figure1},
representing a current compromise in geophysics \cite{Yilmaz2001}.

This being posed, what could CGAN bring compared to an $L_p$ loss in the deterministic case?
Let us first discuss the CGAN foundations in the general non-deterministic case,
from the Wasserstein point of view and complementarily to Ref. \cite{Fabbri2017},
and then analyze the deterministic limit.

\section{Wasserstein CGAN for processing sequences}
\label{sec:Wasserstein}

\subsection{Non-deterministic prediction case}
\label{sec:PAN}

For notational purposes, let us consider the ``parameterized'' conditional pdf ${P}_{Y|X}^{(par)}$
whose realizations correspond to the ones of $G^Z_\theta(X)$.
In other words, for any function $D$, ${P}_{Y|X}^{(par)}$ is defined so that
\begin{eqnarray}
\label{eq:pdf_G}
\mathbb{E}_{Y^{(par)}\sim P_{Y|X}^{(par)}} D(Y^{(par)})
=
\mathbb{E}_{Z\sim P_Z} D(G^Z_\theta(X))
,
\end{eqnarray}
where we keep the superscript $^{(par)}$ to make explicit which random variable is related to the parameterized pdf.
Note that imposing a Gaussian parameterization to ${P}^{(par)}_{{Y}|{X}}$ and using cross-entropy (XE) as a loss
leads to eq. (\ref{eq:dist3}), as recalled in Appendix \ref{app:XE}.
This allows us to understand from another point of view the conclusions of \S \ref{sec:more_det}:
$L_p$ losses are suited when the outcomes follow Gaussian statistics,
and are not suited when the Gaussian assumption is too simplistic (as often with structured outcomes).

We now wish to define a similarity measure between the two joint pdfs
${P}_{Y,X}$ and ${P}_{Y,X}^{(par)}=P_X{P}_{Y|X}^{(par)}$,
without having to consider any 
parameterization like the Gaussian one.
XE is not adapted and Wasserstein distances \cite{Villani2003} seem like a natural choice.
We propose the following Wasserstein-based formulation suited to joint pdfs with same marginal ($p\ge 1$ and $r\ge 1$):
%
\begin{eqnarray}
\label{eq:W1_pred}
JW_{L_p}({P}_{Y,X},P_{Y,X}^{(par)})
&=&
\mathbb{E}_{X\sim {P}_{X}}
W_{L_p}({P}_{Y|X},P_{Y|X}^{(par)})
\\
W_{L_p}({P}_{Y|X},P_{Y|X}^{(par)})
&=&
\Big(
\inf_{{\Pi}^X_{Y,Y^{(par)}}}
\mathbb{E}_{({Y}, Y^{(par)})\sim {\Pi}^X_{{Y}, Y^{(par)}}}
||{Y}- Y^{(par)}||_{L_p}^r
\Big)^\frac{1}{r}
\nonumber
.
\end{eqnarray}
%
${P}_{Y|X}$ and $ {P}_{Y|X}^{(par)}$ 
are considered as single parameter pdfs for each realization of $X$,
and  $W_{L_p}$ represents a $r$-Wasserstein distance between ${P}_{Y|X}$ and $P_{Y|X}^{(par)}$ for any $L_p$-norm choice in the output image space \cite{Villani2003}.
The infimum is taken over all joint pdfs
${\Pi}^X_{{Y}, Y^{(par)}}$
with marginals ${P}_{Y|X}$ and $P_{Y|X}^{(par)}$,
$X$ being considered as a parameter.
%
%
Then, the expectation over all realizations of $X$ is taken to obtain $JW_{L_p}$, representing a distance between the joint pdfs ${P}_{Y,X}$ and $P_{Y,X}^{(par)}$,
as demonstrated in Appendix \ref{app:Wasserstein3}.
Switching to the dual formulation and taking $r=1$
allows to simplify the second line of eq. (\ref{eq:W1_pred}) 
into the Kantorovitch-Rubinstein (KR) formulation
(see \cite{Villani2003} section 1.2 and \cite{arjovsky2017wasserstein})
\begin{eqnarray}
\label{eq:W2_pred}
W_{L_p}({P}_{Y|X},P_{Y|X}^{(par)})
&=&
\sup_{||D_X||^{ }_{Lip_{L_p}}\le 1}
\Big[
\mathbb{E}_{Y\sim P_{Y |X}} D_X(Y)
-
\mathbb{E}_{Y^{(par)}\sim P_{Y|X}^{(par)}} D_X(Y^{(par)})
\Big]
.
\end{eqnarray}
%
Note that the ``discrimator'' $D_X:\mathcal{Y}\rightarrow \mathbb{R}$ 
is parameterized by the input data
and is constrained to be 1-Lipschitz for 
the $L_p$-norm, i.e. $||D_X||^{ }_{Lip_{L_p}}\le 1$.
The corresponding ``Lipschitz norm'' is defined by
$
||D_X||^{ }_{Lip_{L_p}}=\sup_{\tilde Y\ne \tilde Y^{(2)}}
\frac{|D_X(\tilde Y)-D_X(\tilde Y^{(2)})|}{||\tilde Y- \tilde Y^{(2)} ||^{ }_{L_p}},
\forall (\tilde Y,\tilde Y^{(2)}) \in \mathcal{Y}\times \mathcal{Y}$
\cite{Villani2003,arjovsky2017wasserstein}.
As demonstrated
in Appendix \ref{app:Wasserstein1},
if $D_X(\tilde Y)$ is differentiable with respect to $\tilde Y$,
the Lipschitz norm simplifies into
\begin{eqnarray}
\label{eq:LipX}
||D_X||^{ }_{Lip_{L_p}}
=
\sup_{\tilde Y\in \mathcal{Y}}
\Big|\Big|
\frac{\partial D_X(\tilde Y)}{\partial \tilde Y}
\Big|\Big|_{L_q}
\quad\text{with}\quad
1/p+1/q=1
.
\end{eqnarray}


Inserting eq. (\ref{eq:pdf_G}) into eq. (\ref{eq:W2_pred}) to come back to $G^Z_\theta$,
we finally obtain the following more tractable form
\begin{eqnarray}
\label{eq:W2_pred2}
JW_{L_p}(G^Z_\theta)
=
\mathbb{E}_{X\sim {P}_{X}}
\sup_{||D_X||^{ }_{Lip_{L_p}}\le 1}
\Big[
\mathbb{E}_{Y\sim P_{Y |X}} D_X(Y)
-
\mathbb{E}_{Z\sim P_Z} D_X(G^Z_\theta(X))
\Big]
.
\end{eqnarray}
Eqs. (\ref{eq:LipX}) and (\ref{eq:W2_pred2})
provide an adversarial training framework for predictive tasks:
$JW_{L_p}$ contains a supremum principle on $D_X$,
but the result is to be minimized with respect to $\theta$ during the training.

The discriminator's
parameterization by the input data $\tilde X\in \mathcal{X}$
establishes the relation with CGAN \cite{Mirza2014,Isola2017} and its Wasserstein counterpart \cite{Fabbri2017},
that is known.
However, we underline some formal points that were not discussed in previous works to our knowledge:
\begin{itemize}[leftmargin=1cm ,parsep=0cm,itemsep=0cm,topsep=0cm]
\item
The discriminator's dependency on the input data can possibly be strong and discontinuous,
leading in the general case to one different discriminator per input data in eq. (\ref{eq:W2_pred2}).
Of course,
this would be inefficient numerically
and is usually not necessary
(especially when images lie in low dimensional manifolds, do not vary rapidly).
However, the considerations in this section lead to some clarification on the possibility of
using a different discriminator architecture per group of input data with similar properties
within W-CGAN if needed.
\item
Eq. (\ref{eq:LipX}) provides the generalization to $L_{p\ne 2}$-norms
to the derivative-based Lipschitz constraint of Ref. \cite{Gulrajani2017}.
\item
We established that $JW_{L_p}$ represents a distance between two joint pdfs with same marginal.
\item
Appendix \ref{app:Wasserstein3b} establishes the link with the scheme obtained starting from
a Wasserstein distance between joint pdfs without same marginals \cite{Courty2017}.
\end{itemize}

We mentioned in \S \ref{sec:more_det} that training with an $L_p$ loss,
eq. (\ref{eq:dist3}),
compares one realization of $P_{Y |X}$ to one $Z$-realization of $G^Z_\theta(X)$,
which leads to ``averaging''.
Training with $JW_{L_p}$,
eq. (\ref{eq:W2_pred2}),
compares all realizations of $P_{Y |X}$
to all $Z$-realizations of $G^Z_\theta(X)$,
i.e. $G^Z_\theta$ can learn to mimic the realizations of $P_{Y |X}$
and no ``averaging'' occurs.
This fundamental difference would help to produce unblurred results in the non-deterministic case,
when multiple outcomes are related to structured events.
%
In the deterministic prediction case,
however, what $JW_{L_p}$ would bring compared to an $L_p$ loss is unclear.
This is what we discuss now.

\subsection{Advantages of Wasserstein CGAN in the deterministic case?}
\label{sec:PAN_det}

To take the deterministic prediction limit, we use the method mentioned in \S \ref{sec:Notations}.
Eq. (\ref{eq:W2_pred2}) becomes
\begin{eqnarray}
\label{eq:W2_pred2_determ}
JW_{L_p}(G_\theta)
=
\sum_{i=1}^{N_D}
\sup_{||D_{\tilde X_i}||^{ }_{Lip_{L_p}}\le 1}
\Big[
D_{\tilde X_i}(\tilde Y_i)
-
D_{\tilde X_i}(G_\theta(\tilde X_i))
\Big]
,
\end{eqnarray}
where $\tilde X_i$ and $\tilde Y_i$ denote input and output data pairs.
%
%
%
We use a discriminator architecture form
\begin{eqnarray}
\label{eq:discrim_architec}
&&
D_{\tilde X_i}(\tilde Y)
=
\int_{\Omega}
F(\tilde X_i,\tilde Y)(y)
d\mu(y)
,
\end{eqnarray}
where $F(\tilde X_i,\tilde Y)\in\mathcal{Y}$
is parameterized by a convolutional DNN without striding
and an additional last ``layer'' simply represents a sum over the ``pixels''.
The last layer is equivalent to a global average pooling \cite{Isola2017, Radford2016UnsupervisedRL} and
has the advantage 
to make the discriminator DNN model independent of the size of the data
(i.e. the model can be used for any data size).
This architecture allows for an interpretation of what the discriminator learns since $F(\tilde X_i,\tilde Y)$ lies in the output image space.
Also, in our tests on geophysical processing tasks, it led to the highest Wasserstein distance estimates (or supremum values) when training the discriminator.
So, it is the architecture we choose.


Just to gain insight, we first consider a linear parameterization
$F(\tilde X_i,\tilde Y)(y)=\alpha(\tilde X_i)(y)\tilde Y(y)$,
where $\alpha$ is a function of $\tilde X_i$ parameterized by a DNN.
Inserting this in eqs. (\ref{eq:LipX}) and (\ref{eq:W2_pred2_determ}), we obtain\footnote{
We have $JW_{L_p}
\rightarrow
\sum_{i=1}^{N_D}
\sup_{||\alpha(\tilde X_i)||_{L_q}\le 1}
\int_{\Omega}
\alpha(\tilde X_i)(y)
\Big(
\tilde Y_i(y)
-
G_\theta(\tilde X_i)(y)
\Big)
d\mu(y)$,
that necessarily leads 
to the supremum argument
$\alpha^{sup}(\tilde X_i)(y)=|\alpha^{sup}(\tilde X_i)(y)|\times\text{sign}\Big(\tilde Y_i(y)-G_\theta(\tilde X_i)(y)\Big)$ and a saturation of the constraint.
Note that the dependency of $\alpha$ on $\tilde X_i$ is sufficient to define the sign as only one $\tilde Y_i$ is associated to $\tilde X_i$ in the deterministic case.
\label{eq:foot-lin}
}
\begin{eqnarray}
\label{eq:W2_ours_pred3-simple}
\widehat{JW}_{L_p}(G_\theta)
=
\sum_{i=1}^{N_D}
\sup_{||\alpha(\tilde X_i)||_{L_q}= 1}
\int_{\Omega}
|\alpha(\tilde X_i)(y)|\times
\Big|
\tilde Y_i(y) - G_\theta(\tilde X_i)(y)
\Big|
d\mu(y)
.
\end{eqnarray}
In the deterministic prediction limit,
the $L_p$-based loss
defined by eqs. (\ref{eq:dist3}) and (\ref{eq:norm_pixels}) becomes
\begin{eqnarray}
C^p(G_\theta)
&=&
\sum_{i=1}^{N_D}
\int_{\Omega}
\Big|
\tilde Y_i(y) - G_\theta(\tilde X_i)(y)
\Big|^p
d\mu(y)
.
\label{eq:Lp_emppdf2}
\end{eqnarray}
Compared to $C^1$, i.e. eq. (\ref{eq:Lp_emppdf2}) with $p=1$,
we observe that $\widehat{JW}_{L_p}$ adds learnt positive weights $|\alpha(\tilde X_i)|$
with unit $L_q$ norm.
In other words, with a linear parameterization for $F$ in the deterministic case,
W-CGAN ``only'' adds automatic learning of optimal data-dependent variance-like weights
compared to the $L_1$ loss.
In this simple case, these weights can be demonstrated to lead to\footnote{
By definition of the ``dual norm'' (\cite{Brezis1983} chapter I, \cite{Rudin1991} chapter IV),
applied to the first equation in footnote \ref{eq:foot-lin}.
}:
$\widehat{JW}_{L_p}=( C^p )^{1/p}$,
i.e. to a $\widehat{JW}_{L_p}$ that is equivalent to the $L_p$-based loss.
The takaway is that W-CGAN, eq. (\ref{eq:W2_pred2_determ}), should ``at least'' learn
an $L_p$-based loss or reweighting.


%

What about more involved parameterizations for $F$ in eq. (\ref{eq:discrim_architec}), using for instance a convolutional DNN and non-linear activations?
This will produce more involved transformations of $\tilde Y_i$ and $G_\theta(\tilde X_i)$ than a simple reweighting.
Indeed, $\tilde Y_i\rightarrow F(\tilde X_i,\tilde Y_i)$ and $G_\theta(\tilde X_i)\rightarrow F(\tilde X_i,G_\theta(\tilde X_i))$ would then
correspond to a postprocessing of the outputs.
The supremum principle in eq. (\ref{eq:W2_pred2_determ}) allows to learn the postprocessing that makes $JW_{L_p}$ the most sensitive to the differences between the prediction and the output data, i.e. that should
concentrate on the less matched events.
In configurations where adding such a postprocessing would not affect the relative ``positions'' of most of the minimums in the loss valley,
the main effect should be to improve the training convergence and
deal better with the amplitude and noise present in the output data.
These situations should tend to occur amongst others when the postprocessings
do not dramatically affect the gross data amplitudes hierarchy.
This is a first element to interpret when W-CGAN trainings
on deterministic geophysical processing sequences may not help to produce a real improvement.
The case of Fig.~\ref{fig:figure2} possibly falls in this category,
that will be further discussed in \S \ref{sec:CCGAN_results}.

Another element is that the method contains free parameters.
Two of these are related to the Lipschitz constraint:
$q$ (or $p$) in eq. (\ref{eq:LipX}), for which $q\approx 1$ represented a good value, and a weight
to impose the constraint using for instance the method of Ref.~\cite{Gulrajani2017}.
Also, like in Refs.~\cite{Ledig2017,Halpert2018,Zhang2020}, we observed
$JW_{L_p}$ has to be combined to an $L_{p'}$ loss to give correct results,
thus an additional weight is needed.
We chose $p'=1.5$ and tuned the weight so that $JW_{L_p}$ and $L_{1.5}$ losses contribute equally,
to obtain the results in Fig.~\ref{fig:figure2}.
As these hyper-parameters are data dependent,
it may explain why W-CGAN did not produce a systematic improvement in our tests.
The question of tuning the parameters  the best for any kind of data is important but goes beyond the scope of this paper and is left for a future study.

\subsection{Content CGAN: An adversarial way to train a content loss}
\label{sec:CCGAN}

Another difficulty with W-CGAN is that it is not feasible to resolve exactly the supremum principle
in $JW_{L_p}$, eq. (\ref{eq:W2_pred2_determ}), at each iteration.
This can lead to slowness in the training and possibly sometimes to ``localized'' unstabilities,
that may contribute to the explanation of some poor results.
We propose to tackle a part of this specific problem by
a heuristic reformulation of eq. (\ref{eq:W2_pred2_determ}).
Note that the linearized case result of \S \ref{sec:PAN_det}
can equivalently be recovered by firstly imposing the following form to $F$
\begin{eqnarray}
\label{eq:W2_content_000}
F(\tilde X_i,\tilde Y)(y)\rightarrow F(\tilde X_i,\tilde Y)(y)\times\text{sign}\Big( F(\tilde X_i,\tilde Y_i)(y)-F(\tilde X_i,G_\theta(\tilde X_i))(y)\Big)
,
\end{eqnarray}
and then do the linearized approximation
(remember footnote \ref{eq:foot-lin} and beware that the argument of the $\text{sign}$ does not depend on $\tilde Y$ but on $\tilde Y_i$, which is important for the Lipschitz norm, eq. (\ref{eq:LipX})).
Keeping this form for 
any non-linear parameterization of $F(\tilde X_i,\tilde Y)$ and inserting eq. (\ref{eq:W2_content_000}) in eq. (\ref{eq:W2_pred2_determ}), we obtain~
\begin{eqnarray}
\label{eq:W2_content}
\overline{JW}_{L_p}(G_\theta)
=
\sum_{i=1}^{N_D}
\sup_{
||D_{\tilde X_i}||^{ }_{Lip_{L_p}}
\le 1
}
\int_{\Omega}
\Big|
F(\tilde X_i,\tilde Y_i)(y) - F(\tilde X_i,G_\theta(\tilde X_i))(y)
\Big|
d\mu(y)
,
\end{eqnarray}
where $D_{\tilde X_i}$ is defined through eqs. (\ref{eq:discrim_architec}) and (\ref{eq:W2_content_000}).
Eq. (\ref{eq:W2_content}) looks like an $L_1$-based content loss 
\cite{Ledig2017}
that would adversarially be trained, simultaneously with the $G_\theta$ training.
A good content loss should tend to maximize the differences between a prediction that has been ``postprocessed'' (through the DNN $F$) and the corresponding similarly postprocessed output data.
This is achieved by the supremum principle in eq. (\ref{eq:W2_content}), where the Lipschitz constraint provides robustness (to avoid singularities...).
%
This heuristical reasoning leads to our ``Content CGAN'' (C-CGAN) loss.
%
%
Among other advantages,
the C-CGAN loss always remains positive, even if the supremum principle is not well resolved at some iteration,
whereas the W-CGAN loss, eq. (\ref{eq:W2_pred2_determ}), might not.

\subsection{DNN architectures and results}
\label{sec:CCGAN_results}

In the results presented in Fig.~\ref{fig:figure2},
the DNN inputs a ghosted image $\tilde X$ and predicts a deghosted image $G_\theta(\tilde X)$.
The prediction function $G_\theta$ architecture is Unet inspired \cite{Ronneberger2015}.
The $F$ function architecture, that defines the discriminator through eq. (\ref{eq:discrim_architec}), is Denet inspired \cite{Remez2017}.
For the $\tilde X$-dependency of $F$, we found it sufficient to concatenate $\tilde X$ to the first Denet layer;
however, this certainly would deserve a specific study for the reason underlined in \S \ref{sec:more_det}.
As mentioned in \S \ref{sec:PAN_det}, we used $q\approx 1$ in the Lipschitz norm, eq. (\ref{eq:LipX}), and 
combined W-CGAN (or C-CGAN) to the $L_{1.5}$ loss, so that both contribute equally.
Fig.~\ref{fig:figure2} shows a result at 40 epochs, 
before full convergence (150 epochs).
C-CGAN gave better results than CGAN or $L_{1.5}$ only, on our data.
However, the main benefit of C-CGAN is here to accelerate the training.
At full convergence we observed the differences between C-CGAN and $L_{1.5}$ become quite smaller,
possibly for the reason outlined in \S \ref{sec:PAN_det}.

\begin{figure}[h]
\centering
\includegraphics[width=0.96\linewidth]{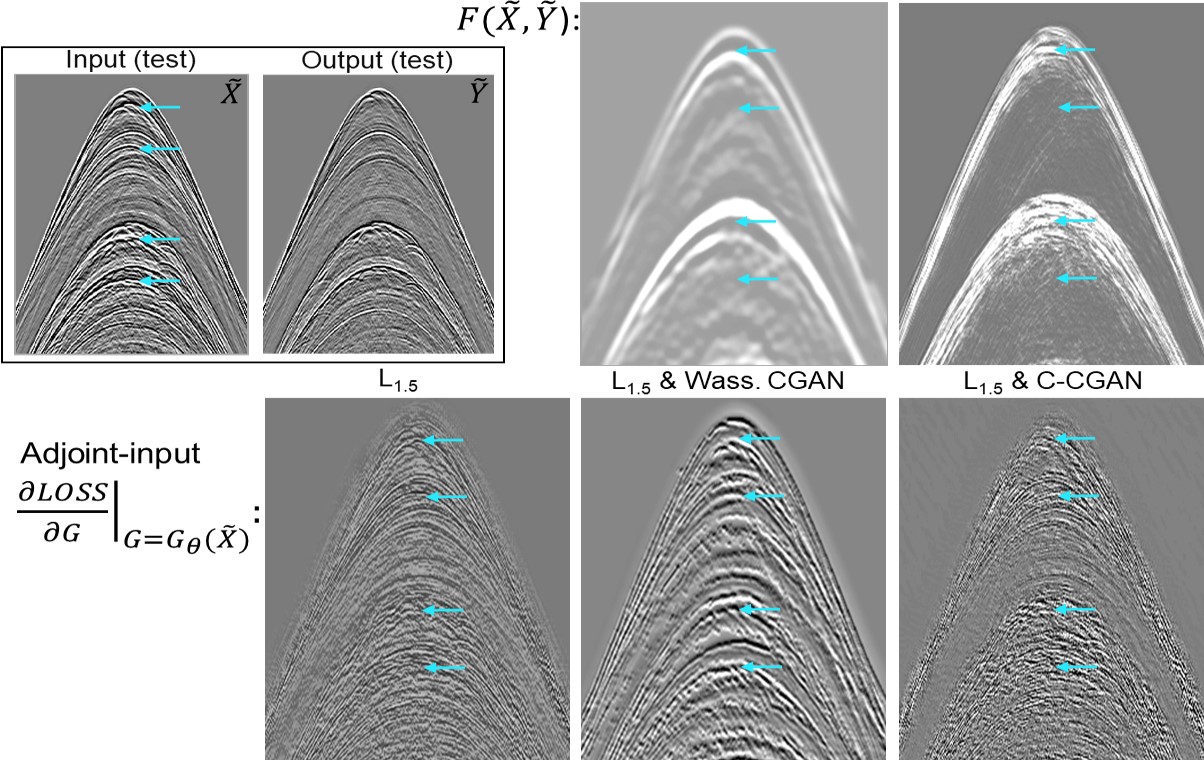}
\caption{
Deghosting task and same data than in Fig.~\ref{fig:figure2}.
This figure illustrates what the discriminator learns
from the $F(\tilde X,\tilde Y)$ (top)
and adjoint-input (bottom) point of views.
We observe that W-CGAN and $L_{1.5}$ tend to ``put a weight'' on many areas, regardless of their relative importance,
while C-CGAN tends to learn to concentrate more on the important areas for deghosting, i.e. around the blue arrows.
}
\label{fig:figure3}
\end{figure}

Fig.~\ref{fig:figure3} 
proposes a way to 
demystify what the discriminator has learnt and interpret the interest of C-CGAN.
Firstly through
$F(\tilde X_i,\tilde Y_i)(y)$, defined in the output image space $\mathcal{Y}$.
The figure shows that C-CGAN learns to concentrate on the most important events, i.e. around the ghosts; this is satisfying and contributes to explain why C-GCAN achieves better deghosting more rapidly.

Secondly, we consider so-called ``adjoint-input'' $\frac{\partial LOSS(G)}{\partial G(y)}\big|_{G=G_\theta(\tilde X_i)}$,
which is back-propagated in $G_\theta$ to compute how to update $\theta$ \cite{Lecun1988}.
The adjoint-input is also defined in the $\mathcal{Y}$ space
and allows for a visualization of the areas where the events should be better predicted after the update.
Fig.~\ref{fig:figure3} shows that the $L_{1.5}$ loss adjoint-input tends to ``put the weight'' on all areas, regardless of their relative importance,
thus not to concentrate more specifically on the ghost areas.
The W-CGAN adjoint-input also tends to concentrate on many areas, contributing to explain why it brought no improvement in Fig.~\ref{fig:figure2}.
The texture of the latter adjoint-input may seem atypical; we verified that the optimization of eq. (\ref{eq:W2_pred2_determ}) and of $G_\theta$ converged without unstability, but further analysis regarding the hyper-parameters mentioned in \S \ref{sec:PAN_det} is on the way.
The C-CGAN adjoint-input, however, learned to concentrate more specifically around
the ghost areas,
contributing to explain why it converges more rapidly towards an acceptable solution in Fig.~\ref{fig:figure2}.
Note, however, that the C-CGAN adjoint-input does not exhibit a very strong change in the amplitudes hierarchy compared to the $L_{1.5}$ adjoint-input,
contributing to explain that the main effect of C-CGAN would here be to improve the training convergence
(remind \S \ref{sec:PAN_det}).
\section{Conclusion and future work}

We proposed a theoretical analysis of the CGAN framework for predictive tasks.
We took the opportunity to establish the CGAN foundations from the Wasserstein point of view,
and pointed out what CGAN should bring compared to an $L_p$ loss in the deterministic prediction case.
We discussed that W-CGAN may perform more poorly than expected
when the corresponding data-space ``postprocessings'' 
would not affect the relative ``positions'' of most of the minimums in the loss valley (for instance when they do not dramatically affect the gross data amplitudes hierarchy),
or due to a difficulty with automatically tuning the corresponding hyper-parameters.
Another difficulty is that the W-CGAN loss represents a distance 
only if the supremum principle is perfectly resolved numerically;
our C-CGAN formalism helps to keep positivity and gave better results on our data.

This first analysis is still to be confirmed by further studies.
It is certainly data dependent (geophysical data being specific, with very structured and continuous events).
Among others, understanding ``physically'' how to tune the CGAN hyperparameters for any kind of data is important in the deterministic as well as in the non-deterministic prediction cases;
this will be a future study.

\section*{Broader Impact}

Learning an efficient representation that mimics involved processing sequences can bring value in a general industrial context, not only in geophysics.
The goal can be to take the best of various existing workflows, increase turnaround or obtain a processing guide.

\begin{ack}
The authors are grateful to CGG and Lundin for the permission to publish this work.
The authors are indebted to Nicolas Salaun, Samuel Gray, 
Thibaut Allemand, Gilles Lambar\'e,
 Mathieu Chambefort and Stephan Cl\'emen\c con for enlightening discussions and collaboration.

All funding of this work by CGG.
\end{ack}

\small
\bibliographystyle{ieeetr}
\bibliography{biblio}
\normalsize

\newpage
\begin{appendix}

\section*{Appendix}

\section{Cross-entropy and Gaussian}
\label{app:XE}

Training  can aim at optimizing a parameterized joint pdf ${P}_{Y,X}^{(par)}=P_X{P}_{Y|X}^{(par)}$ to mimic ${P}_{{Y},{X}}$
the best from a given similarity measure or loss point of view.
Cross-entropy represents a common loss choice to minimize,
Ref. \cite{Goodfellow2016} section 6.2.1.2:
\begin{eqnarray}
\label{eq:XE2}
XE({P}_{Y,X},{P}_{Y,X}^{(par)})
&=&
-\mathbb{E}_{{(Y,X)}\sim {P}_{{Y,X}}}
\ln
\Big(
{P}_{X}(X){P}_{Y|X}^{(par)}(Y|X)
\Big)
.
\end{eqnarray}
%
Dealing with a very general parameterization for ${P}_{Y|X}^{(par)}$ is unfeasible when $\mathcal{Y}$ has a large dimensionality.
However, a simple generalized Gaussian parameterization may be considered 
(the normalization factor is implicit),
see Ref. \cite{Tar05} section 4.3:
\begin{eqnarray}
{P}^{(par)}_{{Y}|{X}}(Y|X)
\propto
\exp\Big( -||Y-G_\theta(X)||_{L_p}^p \Big)
.
\label{eq:condpdf_theta}
\end{eqnarray}
When eq. (\ref{eq:condpdf_theta}) is inserted into eq. (\ref{eq:XE2}) and the terms that do not contribute to the $G_\theta$ optimization are omitted,
we obtain eq. (\ref{eq:dist3}).
Note that ``standard-deviation-like'' weights $\sigma(y)$, with $\infty>1/\sigma(y)>0$, can be added to the distance to control the width of the Gaussian, i.e. $||.||^{ }_{L_p}\rightarrow ||\frac{1}{\sigma}.||^{ }_{L_p}$.
But they do not affect considerations on the maximum-likelihood of the Gaussian.


\section{Wasserstein distance between two joint pdfs with same marginal}
\label{app:Wasserstein3}

We consider pdfs $P^i_{Y,X}$ (indexed by $i$) with same marginal $P_X$,
i.e. $P^i_{Y,X}=P^i_{Y|X}P_X$.
\begin{eqnarray}
\label{eq:app_PAN_000}
{P}^i_{{Y}|{X}}(\tilde Y|\tilde X)
=
\frac{{P}^i_{{Y},{X}}(\tilde Y,\tilde X)}{{P}_{{X}}(\tilde X)}
\mathbf{1}_{{P}_{{X}}(\tilde X)\ne 0}
,
\end{eqnarray}
where the indicator function 
constrains the support in $\mathcal{X}$ of ${P}^i_{{Y}|{X}}(\tilde Y|\tilde X)$
to be the same as the support of $P_X$.

We here use the notations of \S \ref{sec:PAN}.
$W_{L_p}({P}_{Y|X}^1,P_{Y|X}^{2})$, eq. (\ref{eq:W1_pred}),
represents a Wasserstein distance between the conditional pdfs ${P}_{Y|X}^1$ and $P_{Y|X}^{2}$
for each ``parameter'' $X$.
Knowing this, we check if
\begin{eqnarray}
\label{eq:app_PAN_3}
JW_{L_p}({P}^1_{Y,X},P_{Y,X}^2)
&=&
\mathbb{E}_{X\sim {P}_{X}}
W_{L_p}({P}_{Y|X}^1,P_{Y|X}^{2})
\end{eqnarray}
represents a distance between the joint pdfs ${P}_{Y,X}^1$ and $P_{Y,X}^{2}$,
i.e. if it satisfies the symmetry and separation properties, and the triangle inequality.

$JW_{L_p}$
obviously satisfies the symmetry property:
$JW_{L_p}(P^1_{Y,X},P^2_{Y,X})=JW_{L_p}(P^2_{Y,X},P^1_{Y,X})$.

It also satisfies the separation property
\begin{eqnarray}
\label{eq:app_PAN_2}
&&
{P}^1_{Y,X}=P^2_{Y,X}
\Rightarrow
{P}^1_{Y|X}=P^2_{Y|X}
\Rightarrow
W_{L_p}({P}^1_{Y|X},P^2_{Y |X})=0
\Rightarrow
JW_{L_p}({P}^1_{Y,X},P_{Y,X}^2)=0
\nonumber\\
&&
JW_{L_p}({P}^1_{Y,X},P_{Y,X}^2)=0
\Rightarrow
W_{L_p}({P}^1_{Y|X},P^2_{Y |X})=0
\Rightarrow
{P}^1_{Y|X}=P^2_{Y|X}
\Rightarrow
{P}^1_{Y,X}=P^2_{Y,X}
\nonumber
.
\end{eqnarray}
The only subtlety lies in the first implication of the first line and the first implication of the second line;
they are true because
$P^1_{Y,X},P^2_{Y|X}$ and $P_X$ all have the same support as mentioned above after eq. (\ref{eq:app_PAN_000}).

Finally, let us check the triangle inequality. As $W_{L_p}$ represents a distance, we have
$W_{L_p}({P}^1_{Y|X},P^3_{Y |X}) \leq W_{L_p}({P}^1_{Y|X},P^2_{Y |X}) + W_{L_p}({P}^2_{Y|X},P^3_{Y |X})$.
Taking the $P_X$ expected value straightforwardly leads to
\begin{eqnarray}
\label{eq:app_PAN_1}
JW_{L_p}({P}^1_{Y,X},P_{Y,X}^3)
&\leq&
JW_{L_p}({P}^1_{Y,X},P_{Y,X}^2) + JW_{L_p}({P}^2_{Y,X},P_{Y,X}^3)
\nonumber
.
\end{eqnarray}
Thus $JW_{L_p}$ defines a distance between two joint pdfs with same marginal.

\section{Lipschitz norm dual representation}
\label{app:Wasserstein1}

We consider a real Banach space $\mathcal{Y}$ 
indexed by the positions $y$
in a set $\Omega$ that is measurable for the measure $\mu(y)$.
We denote by
$||.||^{ }_{d}$ the corresponding norm,
not necessarilly an $L_p$ norm.

Considering a function $D:\mathcal{Y}\rightarrow \mathbb{R}$,
the Lipschitz semi-norm is defined by
\begin{eqnarray}
\label{eq:W0_app00}
||D||^{ }_{Lip_{d}}=\sup_{\tilde Y\ne \tilde Y^{(2)}}
\frac{|D(\tilde Y)-D(\tilde Y^{(2)})|}{||\tilde Y- \tilde Y^{(2)} ||^{ }_{d}}
,
\quad
\forall (\tilde Y,\tilde Y^{(2)}) \in \mathcal{Y}\times \mathcal{Y}
.
\end{eqnarray}
Identifying constant functions, we obtain a norm;
this is implicit in the article, hence the use of ``Lipschitz norm'' throughout the text.

Imposing Lipschitzity through eq. (\ref{eq:W0_app00}) is unpractical numerically,
as a global property of $D$ is considered.
We would like to turn this global property into a local neighborhood one.
Taking $\tilde Y^{(2)}=\tilde Y + \tilde \Upsilon$ with $\tilde \Upsilon\in \mathcal{Y}$
and choosing a $D(\tilde Y)$ that is derivable with respect to $\tilde Y$, the Lipschitz norm can be rewritten\footnote{
The second equality is obtained by proving two inequalities. The first inequality involves computing the  derivative of $D(\tilde Y)$ around $\tilde Y$. The second inequality involves integrating the derivative of $D(\tilde Y)$ from $\tilde Y$ to ${\tilde Y}^{(2)}$.
}
\begin{eqnarray}
\label{eq:W0_app2}
||D||^{ }_{Lip_{d}}
&=&
\sup_{\tilde Y}\sup_{\tilde \Upsilon \ne 0}
\frac{|D(\tilde Y + \tilde \Upsilon)-D(\tilde Y)|}{||\tilde\Upsilon ||^{ }_{d}}
=
\sup_{\tilde Y}
\sup_{\tilde \Upsilon\ne 0}
\frac{
\Big|\Big(\frac{\partial D(\tilde Y)}{\partial \tilde Y},\tilde \Upsilon\Big)_\Omega\Big|
}{
||\tilde \Upsilon||^{ }_{d}
}
\nonumber\\
&=&
\sup_{\tilde Y}
\sup_{||\tilde \Upsilon||^{ }_{d}\le 1}
\Big|\Big(\frac{\partial D(\tilde Y)}{\partial \tilde Y},\tilde \Upsilon\Big)_\Omega\Big|
,
\end{eqnarray}
%
%
where the differential is defined through
%
\begin{eqnarray}
\label{eq:W0_app3}
\Big(\frac{\partial D(\tilde Y)}{\partial \tilde Y},\tilde \Upsilon\Big)_\Omega
=
\int_{\Omega}
\frac{\partial D(\tilde Y)}{\partial \tilde Y(y)}
\tilde \Upsilon(y)
d\mu(y)
.
\end{eqnarray}
In the following, $\mathcal{Y}'=\mathcal{L}(\mathcal{Y},\mathbb{R})$ denotes the topological dual of $\mathcal{Y}$,
i.e. the ensemble of the continuous linear functions from $\mathcal{Y}$ into $\mathbb{R}$ ($'$ being the symbol of the topological dual).
Note that the differential takes the form in eq. (\ref{eq:W0_app3}) 
when the Riesz representation theorem (\cite{Brezis1983} chapter IV) can be invoked in $\mathcal{Y}'$.

We introduce the function
$
\Phi_{\frac{\partial D(\tilde Y)}{\partial \tilde Y}}\in\mathcal{Y}'
$
that is ``represented'' by
$\frac{\partial D(\tilde Y)}{\partial \tilde Y}$,
i.e. $
\Phi_{\frac{\partial D(\tilde Y)}{\partial \tilde Y}}(\tilde \Upsilon)
=
\Big(\frac{\partial D(\tilde Y)}{\partial \tilde Y},\tilde \Upsilon\Big)_\Omega
$.
By definition of the ``dual norm'' (\cite{Brezis1983} chapter I, \cite{Rudin1991} chapter IV), we have
$
\Big|\Big|
\Phi_{\frac{\partial D(\tilde Y)}{\partial \tilde Y}}
\Big|\Big|'
=
\sup_{||\tilde \Upsilon||^{ }_{d}\le 1}
\Big|\Big(\frac{\partial D(\tilde Y)}{\partial \tilde Y},\tilde \Upsilon\Big)_\Omega\Big|
=
\Big|\Big|
\frac{\partial D(\tilde Y)}{\partial \tilde Y}
\Big|\Big|_{d'}
$.
In other terms, if the Riesz representation theorem can be invoked in $\mathcal{Y}'$ (\cite{Brezis1983} chapter IV),
the dual norm can be ``represented'' by a norm $||.||^{ }_{d'}$ on $\frac{\partial D(\tilde Y)}{\partial \tilde Y}$.
Combining these results with eq. (\ref{eq:W0_app2}) gives:
\begin{eqnarray}
\label{eq:W4_appZZ}
\text{If a Riesz representation theorem exists for $\mathcal{Y}'$:}
\quad||D||^{ }_{Lip_{d}}
=
\sup_{\tilde Y}
\Big|\Big|
\frac{\partial D(\tilde Y)}{\partial \tilde Y}
\Big|\Big|_{d'}
.
\end{eqnarray}

Eq. (\ref{eq:W4_appZZ}) is true for any norm in finite dimensional Banach spaces,
that is the case when $\Omega$ represents a finite pixels grid and $\mu(y)$ the counting measure.
Then, a straightforward choice for the norm $d$ is the $L_p$ norm, 
eq. (\ref{eq:norm_pixels}), for which we have (\cite{Brezis1983} chapter IV)
\begin{eqnarray}
\label{eq:dual_p}
&&
||.||^{ }_{d}=||\frac{1}{\sigma}.||^{ }_{L_p}
\quad\Rightarrow\quad
||.||^{ }_{d'}
=
||\sigma.||^{ }_{L_q}
\quad\text{with}\quad
1/p+1/q=1,
\quad p\ge 1
.
\end{eqnarray}
For completeness, we introduced ``standard-deviation-like'' weights $\sigma(y)$,
that must satisfy $\infty>1/\sigma(y)>\epsilon$ where $\epsilon >0$,
even if in our applications $\sigma(y)=1$ is taken.
Inserting eq. (\ref{eq:dual_p})  in eq. (\ref{eq:W4_appZZ}) leads to
%
\begin{eqnarray}
||D||^{ }_{Lip_{L_p}}
&=&
\sup_{\tilde Y\in \mathcal{Y}}
\Big(
\int_{\Omega} 
\sigma(y)^q
\Big|\frac{\partial D(\tilde Y)}{\partial \tilde Y(y)}\Big|^q d\mu(y)
\Big)^\frac{1}{q}
\nonumber\\
&=&
\sup_{\tilde Y\in \mathcal{Y}}
\Big|\Big|
\sigma\frac{\partial D_X(\tilde Y)}{\partial \tilde Y}
\Big|\Big|_{L_q}
\quad\text{with}\quad
1/p+1/q=1,
\quad p\ge 1
.
\label{eq:dual_p2}
\end{eqnarray}
So, if $\mathcal{Y}$ represents a finite dimensional $L^p(\Omega)$ space,
the ``naturally'' associated Lipschitz norm is defined by eq. (\ref{eq:dual_p2})
(even if all norms are equivalent in finite dimensional spaces, the ``naturally'' associated dual norm
may lead to better numerical convergence properties...).

What if $\mathcal{Y}$ represents an infinite dimensional $L^p(\Omega)$ space
(that is the case for instance when $\Omega$ represents a continuous set and $\mu(y)$ the Lebesgue measure)?
Eqs. (\ref{eq:dual_p})-(\ref{eq:dual_p2}) remain valid but only for $\infty > p \ge 1$.
Indeed, 
no Riesz representation theorem then exists for $\mathcal{Y}'$ when $\mathcal{Y}=L^{\infty}(\Omega)$.

\section{General formulation of the Wasserstein distance between two joint pdfs}
\label{app:Wasserstein3b}

We here consider two joint pdfs ${P}_{Y,X}=P_X{P}_{Y|X}$ and ${P}_{Y,X}^{(2)}=P^{(2)}_X{P}_{Y|X}^{(2)}$
that do not necessarilly have the same marginals, i.e. $P_X\ne P^{(2)}_X$ in the general case.
We consider the $\mathcal{Y}\times\mathcal{X}$ space to be a Banach space
and denote the corresponding norm by
$
||(\tilde Y,\tilde X)-(\tilde Y^{(2)},\tilde X^{(2)})||
$,
$
\forall ((\tilde Y,\tilde X),(\tilde Y^{(2)},\tilde X^{(2)}))\in (\mathcal{Y}\times\mathcal{X})\times(\mathcal{Y}\times\mathcal{X})
$.

A $r$-Wasserstein distance between ${P}_{Y,X}$ and ${P}_{Y,X}^{(2)}$ can be defined by ($r\ge 1$)
\cite{Courty2017}
\begin{align}
&JW2({P}_{Y,X},P_{Y,X}^{(2)}) \nonumber
=\\&
\Big(
\inf_{{\Pi}_{({Y},X), (Y^{(2)},X^{(2)})}}
\mathbb{E}_{
(({Y},X), (Y^{(2)},X^{(2)}))
\sim {\Pi}_{({Y},X), (Y^{(2)},X^{(2)})}
}
||(Y,X) - (Y^{(2)},X^{(2)}) ||^r
\Big)^{\frac{1}{r}}
,
\label{eq:Wasserstein3b_1}
\end{align}
where the infimum is taken over all joint pdfs
${\Pi}_{({Y},X), (Y^{(2)},X^{(2)})}$
with marginals ${P}_{Y,X}$ and $P_{Y,X}^{(2)}$.
Switching to the dual formulation for $r=1$, in the same spirit than in \S \ref{sec:PAN},
leads to
\begin{eqnarray}
\label{eq:Wasserstein3b_10}
JW2({P}_{Y,X},P_{Y,X}^{(2)})
&=&
\sup_{||D||^{ }_{Lip2}\le 1}
\Big[
\mathbb{E}_{(X,Y)\sim {P}_{Y,X}} D(Y,X)
-
\mathbb{E}_{(X^{(2)},Y^{(2)})\sim {P}^{(2)}_{Y,X}} D(Y^{(2)},X^{(2)})
\Big]
,
\nonumber
\\
\end{eqnarray}
where the discrimator $D:\mathcal{Y}\times\mathcal{X}\rightarrow \mathbb{R}$ 
is constrained to be 1-Lipschitz for the Lipschitz norm:
\begin{eqnarray}
\label{eq:Wasserstein3b_100}
||D||^{ }_{Lip2}
=
\sup_{(\tilde Y,\tilde X) \ne (\tilde Y^{(2)},\tilde X^{(2)})}
\frac{|D(\tilde Y,\tilde X)-D(\tilde Y^{(2)},\tilde X^{(2)})|}{||(\tilde Y,\tilde X)- (\tilde Y^{(2)},\tilde X^{(2)}) ||}
.
\end{eqnarray}

Now, we consider ${P}_{Y,X}$ and ${P}_{Y,X}^{(2)}$ that have the same marginal,
i.e. $P_X= P^{(2)}_X$. Eq. (\ref{eq:Wasserstein3b_10}) becomes
\begin{eqnarray}
JW2({P}_{Y,X},P_{Y,X}^{(2)})
&\rightarrow&
\sup_{||D||^{ }_{Lip2}\le 1}
\mathbb{E}_{X\sim {P}_{X}}
\Big[
\mathbb{E}_{Y\sim {P}_{Y|X}} D(Y,X)
-
\mathbb{E}_{Y^{(2)}\sim {P}^{(2)}_{Y|X}} D(Y^{(2)},X)
\Big]
.
\nonumber
\\
\label{eq:Wasserstein3b_1000}
\end{eqnarray}

In order to establish a relation between eq. (\ref{eq:W2_pred2}), $JW_{L_p}$, and eq. (\ref{eq:Wasserstein3b_1000}), $JW2$,
we specify a form for the norm on $(X,Y)$.
We consider 
$\mathcal{Y}$ to be the Banach space used throughout this article
(i.e. equipped with the $L_p$-norm)
and 
$\mathcal{X}$ to be a Banach space (possibly indexed in another space than $\Omega$) equipped
with a $||\frac{1}{\sigma}.||^{ }_{L_s}$ norm
that includes a ``standard-deviation-like'' weight like in eq. (\ref{eq:dual_p}).
We pose
\begin{align}
\Vert (\tilde Y,\tilde X)- (\tilde Y^{(2)},\tilde X^{(2)}) \Vert 
= 
\sqrt{ \Vert \tilde Y - \tilde Y^{(2)} \Vert_{L_p}^2 + \Vert \frac{1}{\sigma} (\tilde X - \tilde X^{(2)}) \Vert_{L_s}^2}
.
\label{eq:Wasserstein3b_1001}
\end{align}
With a similar reasonment than in Appendix \ref{app:Wasserstein1} and
choosing a $D(\tilde Y,\tilde X)$ that is derivable with respect to $\tilde Y$ and $\tilde X$,
one can prove that the Lipschitz norm can be represented by\footnote{
The only subtlety is the use of an Euclidean norm between the $\tilde Y$ and $\tilde X$ components. We did this choice to simplify the computation, the dual representation of the Euclidean norm remainaing the Euclidean norm.
}
\begin{eqnarray}
&&
||D||^{ }_{Lip2}
=
\sup_{\tilde Y,\tilde X}
\sqrt{
\Big|\Big|
\frac{\partial D(\tilde Y,\tilde X)}{\partial \tilde Y}
\Big|\Big|_{L_q}^2
+
\Big|\Big|
\sigma\frac{\partial D(\tilde Y,\tilde X)}{\partial \tilde X}
\Big|\Big|_{L_t}^2
}
\nonumber\\
&&
\hspace{4cm}
\quad\text{with}\quad
1/p+1/q=1
\quad\text{and}\quad
1/s+1/t=1
.
\label{eq:LipX_appD}
\end{eqnarray}
Taking the limit $\sigma\rightarrow 0$
and denoting equivalently $D(\tilde Y,\tilde X)$ or $D_{\tilde X}(\tilde Y)$, eq. (\ref{eq:LipX_appD}) reduces to eq. (\ref{eq:LipX}).
Thus, the $JW_{L_p}$ of \S \ref{sec:PAN} can be considered as a limiting case of the $JW2$ of this section.
Indeed, taking $\sigma \rightarrow 0$ aims at imposing $\tilde X \rightarrow \tilde X^{(2)}$ in eq. (\ref{eq:Wasserstein3b_100}).

\end{appendix}

\end{document}